\documentclass[10pt,twocolumn,letterpaper]{article}

\usepackage{cvpr}

\usepackage{graphicx}
\usepackage{amsmath}
\usepackage{amssymb}
\usepackage{booktabs}

\usepackage[pagebackref,breaklinks,colorlinks]{hyperref}

\usepackage[capitalize]{cleveref}
\crefname{section}{Sec.}{Secs.}
\Crefname{section}{Section}{Sections}
\Crefname{table}{Table}{Tables}
\crefname{table}{Tab.}{Tabs.}

\usepackage{bbm}
\usepackage{bbding}
\usepackage{caption}    
\usepackage{subcaption} 
\usepackage{amssymb}
\usepackage{pifont}
\usepackage{enumitem}
\usepackage{wasysym}
\usepackage{multirow}

\usepackage[table]{xcolor}
\definecolor{mygray}{gray}{0.9}
\definecolor{iconblue}{rgb}{0, 0.60, 0.80}
\definecolor{normalgreen}{rgb}{0.52, 0.79, 0.23}
\definecolor{abnormalorange}{rgb}{0.97, 0.62, 0.32}
\def\confYear{2024}

\begin{document}

\title{Video Anomaly Detection and Explanation via Large Language Models}

\author{Hui Lv$^1$, Qianru Sun$^1$\thanks{Corresponding author}\\
$^1$Singapore Management University\\
{\tt\small $^1$\{huilyu, qianrusun\}@smu.edu.sg}
}

\maketitle
\begin{abstract}
Video Anomaly Detection (VAD) aims to localize abnormal events on the timeline of long-range surveillance videos.
Anomaly-scoring-based methods have been prevailing for years but suffer from the high complexity of thresholding and low explanability of detection results. 
%
%
In this paper, we conduct pioneer research on equipping video-based large language models (VLLMs) in the framework of VAD, making the VAD model free from thresholds and able to explain the reasons for the detected anomalies.
We introduce a novel network module Long-Term Context (LTC) to mitigate the incapability of VLLMs in long-range context modeling.
We design a three-phase training method to improve the efficiency of fine-tuning VLLMs by substantially minimizing the requirements for VAD data and lowering the costs of annotating instruction-tuning data.
%
%
%
Our trained model achieves the top performance on the anomaly videos of the UCF-Crime and TAD benchmarks, with the AUC improvements of +3.86\% and +4.96\%, respectively.
More impressively, our approach can provide textual explanations for detected anomalies. \emph{Our code is in the Appendix.}
\end{abstract}
\section{Introduction}
\label{sec:intro}
Video Anomaly Detection (VAD) 
is to identify
unexpected events in video sequences. 
It has practical applications that span a multitude of fields including
intelligent manufacturing~\cite{Huang2022Survey}, traffic surveillance~\cite{kamijo2000traffic,lv2021localizing} and public security~\cite{mohammadi2016angry,sultani2018real}. 
%
Conventional VAD methods~\cite{lv2021localizing,lv2023unbiased,tian2021weakly,sultani2018real,zhang2023video,zhou2023dual}
are designed to predict anomaly scores along the timeline of the video, i.e., one anomaly score for each video frame.
A higher score indicates a higher possibility of being abnormal on the frame.
%
These anomaly-score-based designs are simple to implement but remain far away from the ideal agents of VAD which should be both \underline{automatic} (i.e., free from manually-selected thresholds) and \underline{explainable} (i.e., being able to explain why an event is abnormal). 

\begin{figure}[t]
    \centering
    \includegraphics[width=0.47\textwidth]{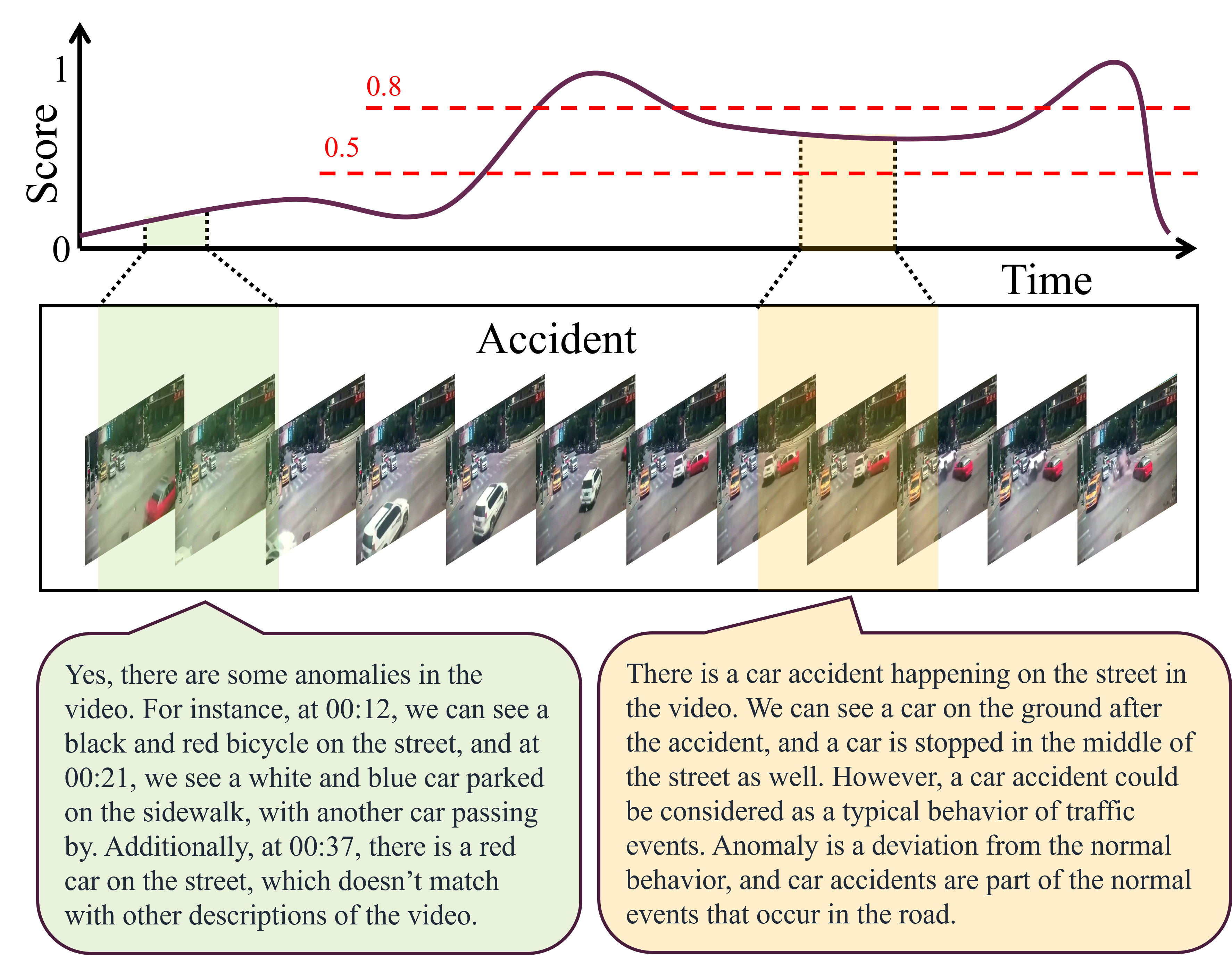}
    \caption{Prediction scores from a baseline VAD model, and clip descriptions by using VLLMs, for a car accident video (as shown in the middle of the figure). 
    On the score curve, the red dashed lines denote anomaly thresholds. 
    The bottom shows the answers from Video-LLaMA~\cite{zhang2023video} by feeding it with two pairs of video clips and questions, respectively: \{\textcolor{normalgreen}{Green}: a normal video clip, ``Is there any anomaly in the video?''\} and \{\textcolor{abnormalorange}{Orange}: an abnormal video clip, ``Is there a car accident? If so, is it an anomaly?''\}}
    \label{fig:Shortcomings}
\end{figure}

Firstly, it is not intuitive how to determine the optimal threshold given diverse video content as well as abnormal events.
For example, as depicted on the top of Figure~\ref{fig:Shortcomings}, 
using different thresholds on the prediction results (scores) of the VAD model yields different detection outcomes.
Secondly, with a carefully selected threshold, anomalies are localized along the timeline based on only scores, and these scores provide little information for users to comprehend the contexts or ascertain the reasons behind the anomalies.
In this paper, we are interested in the VAD model not merely to automatically identify anomalies but also to provide comprehensive textual explanations.
We incorporate the Video-based Large Language Models (VLLMs) into the framework of VAD, Video-LLaMA~\cite{zhang2023video} in our case, and call the method VAD-LLaMA. In the following, we elaborate on the challenges and our solutions.

Well-trained VLLMs (such as Video-ChatGPT~\cite{maaz2023video}, Videochat~\cite{li2023videochat}, and Video-LLaMA~\cite{zhang2023video}) can generate detailed captions for any input video.
%
However, 
there is a discrepancy between VLLMs' and humans' understanding of anomalies. 
As illustrated in Figure~\ref{fig:Shortcomings}, Video-LLaMA~\cite{zhang2023video} identifies several irrelevant objects in the scene as ``anomalies'', while overlooking the car accident---the real anomaly humans care about.
To solve this issue, we propose a novel Video Anomaly Detector (VADor) equipped with the modules from Video-LLaMA~\cite{zhang2023video}, and a new method for co-training VADor and VLLMs without needing a large amount of domain-specific data and labels.

This co-training poses two key challenges.
\textbf{The first challenge} is that open-sourced VLLMs lack long-range context modeling ability.
They are mostly trained on short videos with simple contexts, but the videos of VAD exhibit high context complexities.
For example, WebVid~\cite{bain2021frozen}, commonly used for fine-tuning VLLMs, features an average video length of only 18 seconds, notably shorter compared to the average length of 240 seconds in the VAD dataset UCF-Crime~\cite{sultani2018real}.
In long videos, anomalies really depend on long-range video contexts. For example, identifying a burglary event requires consideration of preceding activities, such as the breaking of windows or doors, even if the event itself only displays moving valuables outside.
%
%
\textbf{The second challenge} is the lack of VAD data and labels.
The widely-used VAD dataset, UCF-Crime~\cite{sultani2018real}, is used for weakly-supervised VAD, as it offers only video-level anomaly annotations, i.e., given a video, it has only a one-hot label indicating normal or abnormal.
%
Therefore, it is not intuitive how to generate text-based instruction data to fine-tune VLLMs.
Besides, VAD datasets have a small scale, e.g., UCF-Crime contains 1.9K training videos, significantly smaller than the VLLM training datasets such as WebVid~\cite{bain2021frozen} containing 10M videos. The fine-tuning of VLLMs on VAD datasets is thus challenging.
To tackle the first challenge,
we introduce a novel Long-Term Context (LTC) module in the VADor.
The key idea is to integrate the long-term normal/abnormal contexts into the video representation.
First, we split a video into multiple clips, and use the video encoder (VE) of Video-LLaMA to extract the features of each clip.
Taking the features as input, VADor can output an anomaly score for each clip.
Based on the lowest (highest) $K$ anomaly scores, we pick corresponding clip features and stack them into a normal (abnormal) list.
The generation of these two lists is implemented as an online operation for each video: every new clip will be immediately evaluated based on its anomaly score to update the lists (or not).
%
Given the ``raw'' features of the next clip, we integrate the current lists of normal and abnormal features by cross-attention and weighted-sum operations, see Sec.~\ref{Med:MT}, i.e., the way we integrate the long-term contexts of the video into the video representation.
To resolve the second challenge, we propose a three-phase training method.
The first phase is to train a baseline VADor, based on which we can easily form a new VAD dataset with each frame ``annotated'' by an anomaly score.
In the second phase,
we co-train VADor and the proposed LTC on the above dataset.
The primary objective here is to incorporate long-term contextual understanding into the LTC and then use it to enhance the video representation of Video-LLaMA in the final phase.
%
The final phase is to fine-tune Video-LLaMA. Based on the above dataset, we manually compose simple textual templates (showcased in Sec.~\ref{Med:MA}) to generate instruction-tuning data, and then use the data to train only the projection layer of Video-LLaMA. We avoid fine-tuning the entire Video-LLaMA due to the limited scale of the VAD dataset. 
Moreover, to prevent overfitting on VAD videos, we incorporate a diverse training sample set, drawing from both the UCF-Crime and WebVid datasets. The latter has been instrumental in the pre-training of Video-LLaMA.
Our method enhances the efficiency of training VAD-LLaMA by substantially minimizing the requirements for VAD data and lowering the costs of creating instruction-tuning data.
During testing, VAD-LLaMA is capable of not only identifying anomalies from the input video but also outputting textual explanations of the reasons for being abnormal. 


Our contributions are thus three-fold. 
1)~A new approach called VAD-LLaMA that introduces VLLMs for tackling the task of VAD. 2)~A novel LTC module that enhances the long video representation ability of existing VLLMs. 3)~A novel three-phase training method for the proposed VAD-LLaMA, by resolving the issues of lacking VAD data and instruction-tuning data.

\section{Related Work}
\label{sec:RW}
\noindent\textbf{Video anomaly detection (VAD)} has been a prominent research area with diverse real-life applications. 
However, it remains a challenging task primarily due to the scarcity of anomalous data and labels. Consequently, researchers often turn to Weakly Supervised Video Anomaly Detection (WSVAD) methods to address the VAD problem. 
These approaches make use of both normal and abnormal training data, relying on weak annotations provided only at the video-level~\cite{sultani2018real}.
Multiple instance learning (MIL) is the mainstream paradigm that uses video-level labels for training snippet-level anomaly detectors~\cite{sultani2018real,he2018anomaly,zhu2019motion,tian2021weakly,li2022self,wu2021learning,zaheer2020claws,zhang2019temporal}. 
Generally, they embrace the two-stage anomaly detection pipeline, which performs anomaly detection upon pre-extracted features.
In particular, Zhong~\etal~\cite{zhong2019graph} considered the WSVAD task as supervised learning under noise labels and they designed an alternate training procedure to enhance the discrimination of action classifiers.
Lv~\etal~\cite{lv2021localizing} focused on anomaly localization and proposed a higher-order context model as well as a margin-based MIL loss.
Li~\etal~\cite{li2022self} proposed multiple sequence learning, where consecutive snippets with high anomaly scores are selected in MIL learning.
More recently, Lv~\etal~\cite{lv2023unbiased} proposed an unbiased MIL framework for removing the context bias.
And they integrated feature representation fine-tuning and anomaly detector learning into an end-to-end training fashion.
In this paper, we follow the end-to-end manner to tackle the WSVAD problem and make the first effort to introduce VLLMs into VAD for endowing the VAD model with the ability of anomaly description.

\noindent\textbf{Video-based large language models (VLLMs)} have demonstrated remarkable language understanding and reasoning abilities, thanks to the ongoing research efforts in exploring the use of LLMs for processing multi-modal problems~\cite{gao2023llama,li2023videochat}.
Bain~\etal~\cite{bain2021frozen} introduced WebVid, a large-scale dataset of short videos with textual descriptions sourced from stock footage sites.
Based on it, Li~\etal~\cite{li2023videochat} improved image encoders, enabling large models to understand visual content in videos.
Su~\etal~\cite{su2023pandagpt} utilized multi-modal encoders to enable large models to understand six modalities.
Zhang~\etal~\cite{zhang2023video} trained fundamental models to comprehend both the visual and auditory content in videos.
In this work, we focus on the visual modal in videos, since most videos in VAD are collected from road surveillance, which falls short of audio signals. 
By integrating the designed VADor with the VLLMs, we propose a novel approach VAD-LLaMA, which is able to not only detect the anomalies but also explain the details of the anomalies.

\section{Method}
\label{sec:Med}
\begin{figure}[t]
    \centering
    \includegraphics[width=0.47\textwidth]{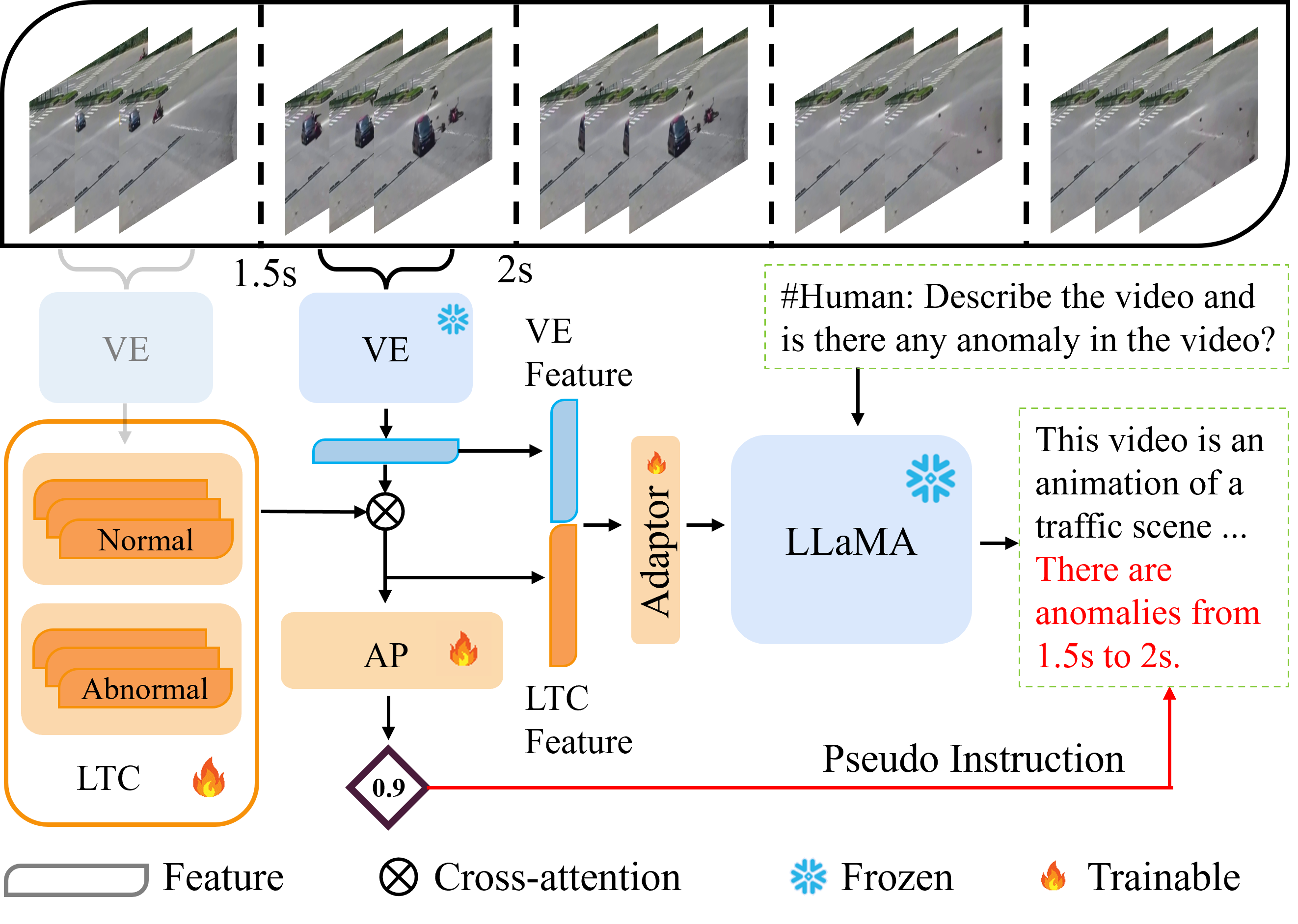}
    \caption{The network architecture of the proposed VAD-LLaMA. It consists of a Video Anomaly Detector (VADor) with the Long-Term Context (LTC) module and a simple Anomaly Predictor (AP), a  projection layer (called Adaptor), and the pre-trained Video-LLaMA~\cite{zhang2023video} (composed by a Video Encoder (VE) and a LLaMA). The training of VAD-LLaMA is decomposed into three phases, and the trainable and frozen modules vary among different training phases. Training phases are given in Figure~\ref{fig:steps}.}
    \label{fig:pipeline}
\end{figure}

\begin{figure*}[t]
    \centering
    \includegraphics[width=1\textwidth]{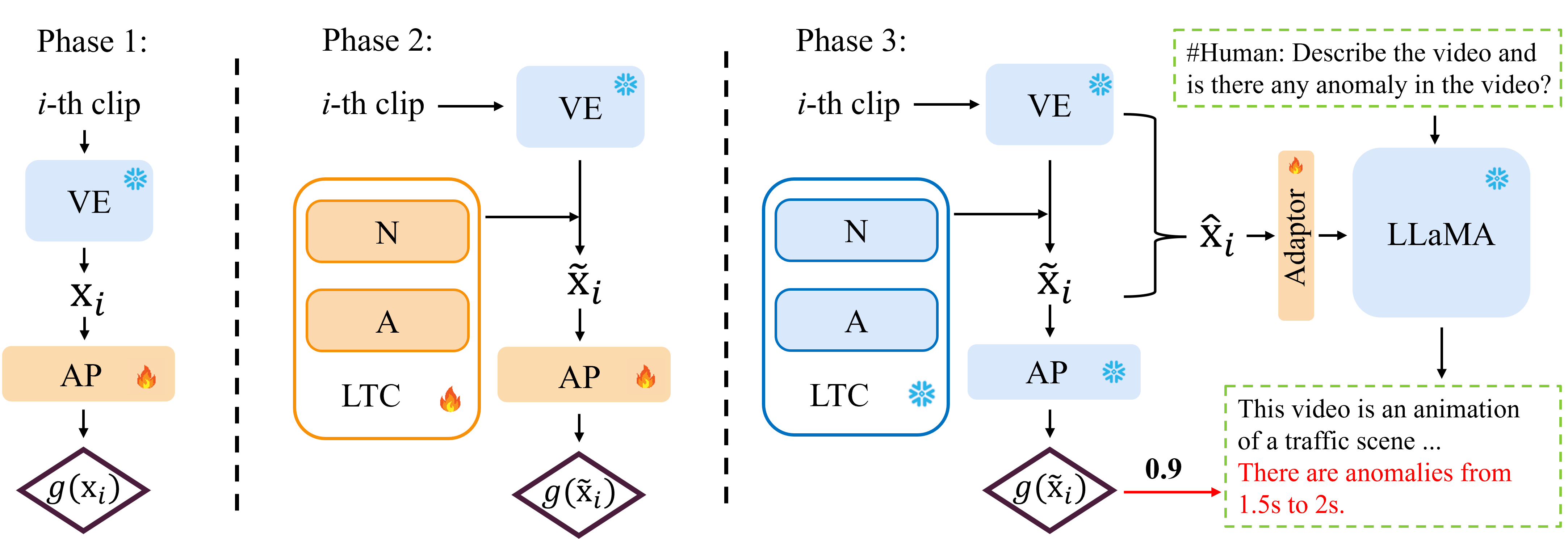}
    \caption{The training phase of VAD-LLMs consists of three phases. 1) VAD baseline training, 2) VAD co-training with LTC, and 3) Instruction-tuning Adaptor. In the LTC module, $\mathbf{N}$ and $\mathbf{A}$ represent the long-term normal and abnormal feature lists, respectively. The red arrow denotes the generation process from anomaly scores to pseudo instructions with text templates.}
    \label{fig:steps}
    \vspace{-4mm}
\end{figure*}

Our VAD-LLaMA architecture is illustrated in Figure~\ref{fig:pipeline} and its training method is shown in Figure~\ref{fig:steps}. It aims to adapt the general video representation knowledge of a pre-trained large video-language model Video-LLaMA~\cite{zhang2023video} to tackle VAD tasks.
Below, we first elaborate on its network architecture in Sec.~\ref{Med:MA}. 
Then, we delve into the specifics of the three-phase training method in Section~\ref{Med:MT}. 

\subsection{Model Architecture}
\label{Med:MA}

\noindent
\textbf{Overview.} As depicted in Figure~\ref{fig:pipeline}, VAD-LLaMA mainly consists of a new VADor, and two pre-trained modules (VE and LLaMA) from Video-LLaMA~\cite{zhang2023video}.
The VADor is built upon the VE and includes a novel LTC module and a simple Anomaly Predictor (AP) $g$ consisting of two fully-connected (fc) layers.
Besides, VAD-LLaMA learns an adaptor $f$ between the VADor and the LLaMA to align their feature distributions.

\noindent
\textbf{VE and Feature Extraction.} 
Given a video sequence, we first divide it into $m$ segments.
For each segment, we randomly sample a video clip (consecutive frames), and feed it into the pre-trained VE to extract clip-level features.
We denote $\mathbf{x}_i, i\in\{1,\ldots, m\}$ as the VE feature of the $i$-th clip.
In this work, the adopted VE from Video-LLaMA~\cite{zhang2023video} consists of an image encoder (BLIP-2~\cite{li2023blip}) and a Video-Qformer, sharing the same architecture with Query Transformer~\cite{li2023blip}. 
The image encoder includes a ViT-G/14 from EVA-CLIP~\cite{fang2023eva} and an image-level Query Transformer.
As aforementioned, these VE features lack long-term context information, as the VE was pre-trained mainly with short and normal videos.


\noindent\textbf{Long-Term Context (LTC) Module}. 
The LTC module is proposed to solve the above challenge.
Specifically, we collect the clip-level VE features with $K$ lowest (highest) anomaly scores and stack them into a normal (abnormal) list. We denote the normal list as $\mathbf{N} = \{\mathbf{n}_j\}_{j=1}^K$, and abnormal list as $\mathbf{A} = \{\mathbf{A}_j\}_{j=1}^K$.
These two lists are online updated and every new clip will be immediately evaluated based on its anomaly score to update the lists (or not).
In addition, we introduce the cross-attention mechanism in the LTC module for integrating the two lists' information into the VE features.
Output features of the LTC module are not only taken as inputs into the AP, but also stacked with the VE features to serve as the visual prompts (input embeddings) of LLaMA.
Based on the LTC-enhanced features, we are able to derive a more robust VADor and also provide comprehensive video contexts for LLaMA.

\noindent\textbf{Feature Adaptor.}
In VAD-LLaMA, the Adaptor $f$ (one fc layer) is added to convert the visual prompts into the same dimension with the inputs of LLaMA and align the visual feature distributions with the pre-trained LLaMA.

\noindent\textbf{LLaMA.}
In this work, we adopt the LLaMA of version vicuna-7b~\cite{zheng2023judging}. 
It's important to highlight that the fine-tuning of LLMs is based on the instruction-tuning data~\cite{bain2021frozen}. Typically, this data is comprised of video instruction pairs, where each pair includes a textual instruction corresponding to the content of the accompanying video. These instructions are often generated using simple templates, commonly in a question-answer format. Here's an illustrative example with the underlined part generated as a pseudo instruction:

\indent\textit{Question: \#\#\# Human: $<$Video$>$ [Video Tokens] $<$/video$>$ [Video Description] Is there any anomaly in the video? } 

\indent\textit{Answer: \#\#\# Assistant: \underline{Yes, there are anomalies from} \underline{1.5s to 2s.}}

In the \textit{Question}, [Video Tokens] denoted the tokens (places) for inserting visual prompts. [Video Description] is simple video clip details, \eg, video length and frame sample rate.
During the instruction-tuning of VAD-LLaMA, the \textit{Question} is first transformed into textual embeddings with a pre-trained LLM (vicuna-7b~\cite{zheng2023judging}) and then concatenated with visual prompts to serve as the inputs of LLaMA.
Later, the textual embeddings transformed from the \textit{Answer} are utilized as the ``ground truth'' of LLaMA's generation.

\subsection{Model Training}
\label{Med:MT}

The training pipeline of VAD-LLaMA is outlined in Figure~\ref{fig:steps}. We implement a three-phase approach. 
In the first phase, clip-level VE features are input into the VADor to establish a baseline for predicting initial anomaly scores. 
In the second phase, these preliminary scores facilitate the aggregation of representative normal and abnormal features within the LTC module. 
This module is co-trained with the VADor to merge long-term contextual information into the process of representation learning and anomaly detection. 
In the third phase, we refine our VAD-LLaMA model by exclusively training the feature adaptor (the projection layer), utilizing the robust features produced by the VE and LTC modules.
%
These features impart a broad understanding of general video content and specific guidance for anomaly detection, which are integral to the instruction-tuning process. Concurrently, the improved anomaly scores, ascertained through the VADor in conjunction with the LTC module, are transformed into pseudo instructions. 
These are then amalgamated with straightforward text templates, serving as the instruction-tuning data for LLaMA.
The details of the training phases are given below.

\noindent\textbf{Phase 1: Training VADor.}
\label{Med:BT}
In this phase, we train a simple VADor baseline by directly passing the VE features through AP $g$, as shown in the left of Figure~\ref{fig:steps}.
Facing the scarcity of anomalous data and labels in VAD, many researchers opt to address the VAD problem in a Weakly Supervised Video Anomaly Detection (WSVAD) framework. 
In this setting, each training video is annotated with a binary anomaly label $y \in \{0, 1\}$ (\ie, normal or abnormal), denoting whether it is categorized as normal or abnormal. 
This allows for training VAD models without the need for frame-level annotations of specific anomalous events, making it more feasible in real-world applications.
In this work, we adopt the same setting as in WSVAD. 
 
The prevailing approach to WSVAD is Multiple Instance Learning (MIL). 
It aims to train a clip-level AP $g$ base on the VE features $\{\mathbf{x}_i\}_{i=1}^m$.
In this process, it distinguishes the most anomalous clip in a normal video (i.e., $y=0$) as normal, and identifies the most anomalous clip within an abnormal video (i.e., $y=1$) as abnormal.
To achieve this, MIL constructs a tuple set $\mathcal{S}$, one tuple for each video, which includes the prediction $y'$ generated by $g$ on the most anomalous snippet and the corresponding video-level label, denoted as $y$. 
This tuple is represented as $(y', y)$, where $y'$ is computed as $\mathrm{max}\{g(\mathbf{x}_i)\}_{i=1}^m$.  The parameters of $g$ are trained by minimizing the binary cross-entropy (BCE) loss:
\begin{equation}
    \textrm{BCE}(\mathcal{S}) = -\mathop{\mathbb{E}}_{(y',y)\sim \mathcal{S}} \left[ y\mathrm{log} (y') + (1-y)\mathrm{log}(1-y') \right].
    \label{eq:1}
\end{equation}
In this way, for a normal video with $y=0$, by \emph{minimizing} $\mathrm{max}\{g(\mathbf{x}_i)\}_{i=1}^m$, $g$ is compelled to assign low abnormal probabilities to all video clips. 
Conversely, for an abnormal video with $y=1$, by \emph{maximizing} $\mathrm{max}\{g(\mathbf{x}_i)\}_{i=1}^m$, $g$ is trained to yield an even higher probability for the most confident abnormal snippet.
Following previous method~\cite{lv2023unbiased} that directly predicts binary logits for the normal and abnormal probabilities.
During the inference of WSVAD, we utilize the abnormal probabilities as anomaly scores to calculate the evaluation metrics (AUC).

\noindent\textbf{Phase 2: Co-Training VADor and LTC.}
\label{Med:CT}
The VADor trained in phase 1 is a MIL-based baseline.
It is trained from VE features.
However, this VE was pre-trained on short videos whose contexts that significantly differ from those in long and complex videos of VAD.

To enhance the long-term video representation ability of VADor, we co-train it with a novel LTC module,
which is designed to encode the most normal as well as the most abnormal events seen in the input video.
As mentioned in Sec.~\ref{Med:MA}, the top-$K$ normal and abnormal VE features are collected and stacked into a normal list $\mathbf{N}$ and an abnormal list $\mathbf{A}$, respectively. 
In the forward pass, the top-$K$ selection is based on the preliminary anomaly scores predicted from the VADor baseline, we pick up the VE features with the $K$ lowest (highest) scores as the items of list $\mathbf{N}$ ($\mathbf{A}$).
Moreover, we online update the lists by re-collecting the historical features based on their anomaly scores, when inputting a new video clip into the LTC module.

Based on the ``memory'' stored in the LTC module, the next question is how to incorporate them into the representation (i.e., the VE features) of new video clips.
To this end, we introduce the cross-attention mechanism to automatically retrieve contextual features from the LTC lists, based on their relevance to the current VE feature.
Specifically, we regard the current VE feature $\mathbf{x}_i$ as the \textbf{query} and the stacked features from LTC lists $\mathbf{N}, \mathbf{A}$ are utilized as the \textbf{key} and \textbf{value} at the same time.
Taking the $i$-th VE feature as an example, the process derives as:
\begin{equation}
    (\mathbf{x}_i \times \mathbf{X}^\top) \times \mathbf{X}, \mathbf{X} \in \{\mathbf{N},\mathbf{A}\},
    \label{eq:2}
\end{equation}
we denote the acquired feature from Eq.~\eqref{eq:2} as $\mathbf{n}_i,\mathbf{a}_i$, separately from $\mathbf{N}$ and $\mathbf{A}$.
Here, $\times$ denotes the dot product. 
In detail, the $i$-th VE feature is first multiplied with the LTC-listed features to generate attention weights, then the relative features are retrieved with these weights,~\ie the higher the feature similarity, the higher the attention weight. 
After that, we combine these features with the VE feature $\mathbf{x}_i$:
\begin{equation}
    \tilde{\mathbf{x}}_i = \mathbf{x}_i + w_n \mathbf{n}_i + w_a \mathbf{a}_i,
    \label{eq:3}
\end{equation}
here, instead of carefully tuning hyper-parameters to manually select a descent weight tuple, we introduce neural soft weights with parameters $w_n, w_a \in W$ to automatically balance the features.
Then the feature after cross-attention is fed into the AP and form a more robust VADor. 
Later, the VADor and the LTC are co-trained and supervised with BCE loss as in ~\eqref{eq:1}.

Note that, to further integrate the short-term historical information that involves the variation of happening events, we add a list for storing the past $K$ VE features, represented as $\mathbf{H} = \{\mathbf{h}_j\}_{j=1}^K$. 
These VE features in the short-term history contribute to learning a more comprehensive feature representation and boost the robustness of VADor.
In this way, we upgrade the LTC module with a plus version, namely Long-Short-Term Context (LSTC) module.
Extensive experiments demonstrate the effectiveness of VADor with the LTC and LSTC module as in Sec.~\ref{sec:Abla}.


\noindent\textbf{Phase 3: Instruction-Tuning Adaptor.}
\label{Med:FT}
In this phase, we incorporate the VADor with the pre-trained VE and LLaMA from Video-LLaMA~\cite{zhang2023video} by adding an adaptor. 
Considering the limited training data in VAD, we opt to freeze the large modules (VADor, VE, and LLaMA) and train only the adaptor that aligns the feature distribution of VADor with the LLaMA.
The frozen modules helps to reduce the model's dependence on the scale of the training data.
Also the features from the well-trained VADor provide a comprehensive understanding of general video content and specific guidance for anomaly detection, which are integral to the instruction-tuning process.

\noindent\textit{Anomaly Prompt}.
To seamlessly incorporate video representations and anomaly information into LLaMA, we utilize the LTC feature $\tilde{\mathbf{x}}_i$ trained in the co-training phase as the \textit{anomaly prompt}. 
As illustrated in Figure~\ref{fig:pipeline}, the anomaly prompt is stacked with the VE feature, resulting in $\hat{\mathbf{x}}_i = [\mathbf{x}_i, \tilde{\mathbf{x}}_i]$. 
Then, we add an adaptor (linear layer) $f$ to project them into the same dimension as the inputs of LLaMA.
The output feature embedding $f(\hat{\mathbf{x}}_i)$ serves as clip-level soft visual prompts that guide the pre-trained LLaMA to generate text that is conditioned on the visual content and anomaly status of the video clip.

\noindent\textit{Pseudo-Instruction.}
There is no frame-level anomaly annotation in WSVAD, so it is not intuitive how to construct temporal instructions for LLaMA.
To address this challenge, 
we propose to convert anomaly scores (output by VADor) into pseudo instructions and manually compose anomaly-related text templates to generate instruction-tuning data.
The process is showcased in the red line of Figure~\ref{fig:pipeline} and Figure~\ref{fig:steps}.

To generate video instruction pairs for VAD data (e.g., UCF-Crime dataset~\cite{sultani2018real}), 
we start by inserting the visual prompts $\{\hat{\mathbf{x}}\}^m_{i=1}$ into the textual embeddings $Q$ of the \textit{Question}. Then, we convert the predicted anomaly score from the VADor into the pseudo instruction as showcased in the \textit{Answer}, \eg, ``0.9'' is transformed into the underlined part of \textit{Answer} in Sec.~\ref{Med:MA}, according to the time duration of the video clip.
Hence, for $i$-th video clip, the video instruction pair becomes $(\hat{y}', \hat{y})$, here $\hat{y}' = [f(\hat{\mathbf{x}}_i), Q]$ stands for the inputs of LLaMA and $\hat{y}$ denotes the textual embeddings transformed from the pseudo instruction in \textit{Answer}.

To prevent overfitting on VAD videos, we incorporate a diverse training
sample set $\mathcal{P}$, drawing from both the UCF-Crime and WebVid datasets.
Finally, we train the adaptor with the cross-entropy loss.
Note that Cross-Entropy (CE) loss is commonly employed for training LLMs~\cite{dou2022empirical}, which quantifies the disparity between the text sequence generated by the model and the target text sequence. The formula of CE loss is derived as follows:
\begin{equation}
    \textrm{CE}(\mathcal{P}) = -\mathop{\mathbb{E}}_{(\hat{y}', \hat{y})\sim \mathcal{P}} \left[ \sum_{j=1}^n {\hat{y}_j} \mathrm{log} ({\textrm{LLaMA}}(\hat{y}')_j) \right],
    \label{eq:4}
\end{equation}
where $n$ is the number of embedding tokens in $\hat{y}$, $\hat{y}_j$ is the true label for token $j$ and ${\textrm{LLaMA}}(\hat{y}')_j$ is the LLaMA-predicted probability for token $j$. 
Additionally, we illustrate the detailed pipeline with a pseudo-code in the Appendix.



\section{Experiments}
\label{sec:Exp}
\subsection{Datasets and Evaluation Metrics}
To verify the performance of our VADor, we conducted extensive experiments and ablations on two standard WSVAD evaluation datasets~\cite{sultani2018real,lv2021localizing}. 
As per the standard in WSVAD, the training videos only have video-level labels, and the test videos have frame-level labels. 
Other details of the experimental setting are given below.

\noindent\textbf{UCF-Crime}~\cite{sultani2018real} is a large-scale dataset comprising 1,900 untrimmed real-world surveillance videos on general scenarios, encompassing both outdoor and indoor environments. 
The dataset boasts a total duration of 128 hours and includes 13 distinct classes of anomalous events.
It is divided according to the standard split, with a training set of 1,610 videos, and a test set of 290 videos.

\noindent\textbf{TAD} dataset~\cite{lv2021localizing} features real-world traffic scene videos, with an average of 1,075 frames per video. These videos encompass over seven common road-related anomaly categories.
The dataset is split into a training set consisting of 400 videos and a test set comprising 100 videos.

\noindent\textbf{Evaluation Metrics}. 
Following previous works~\cite{sultani2018real,lv2023unbiased}, we adopted the Area Under the Curve (AUC) of the frame-level ROC (Receiver Operating Characteristic) as the main WSVAD evaluation metric for TAD and UCF-Crime. 
Intuitively, a larger AUC means a larger margin between the normal and abnormal predictions of video clips, suggesting a superior anomaly classifier.
Taking inspiration from UMIL~\cite{lv2023unbiased}, our evaluation goes beyond calculating AUC for the entire test set, denoted as~$\mathrm{AUC}_O$. 
We also compute the AUC specifically for abnormal videos, referred to as~$\mathrm{AUC}_A$.
It is for excluding normal videos where all clips are normal (label 0), and keeping only the abnormal ones with both kinds of clips (label 0,1). 
This selective evaluation truly challenges a classifier's capability to accurately localize and detect anomalies within a mixed context.
\subsection{Implementation Details}
\label{sec:ID}
Given that VAD videos are predominantly sourced from CCTV surveillance, which typically lacks audio signals, we omit the audio branch while retaining the visual part as the VE for generating video clip features. 
We also implemented the pre-trained LLaMA from Video-LLaMA~\cite{zhang2023video} to retain general video description knowledge.
In our VAD-LLaMA, two fc layers are used to implement the AP $g$, and another two fc layers are used to balance the features as $W$ in the LTC module, each corresponding to a feature list, i.e., normal or abnormal.
All the fc layers in VAD-LLaMA are initialized with random weights and trained to locate and describe the possible anomalies in videos.
The length of LTC lists is uniformly set as 4 by default, according to the ablation study in Sec.~\ref{sec:Abla}. 

We trained our model with the AdamW optimizer~\cite{loshchilov2019decoupled} using an initial learning rate of $1$e-$5$, weight decay of $0.001$, and batch size of $8$ in the training phases 1 and 2.
Due to the large memory consumption of LLaMA, the batch size was reduced to $2$ during phase 3.
We utilized the cosine annealing scheduler and warmed up the learning rate for $5$ epoch among each training phase.
The VADor baseline was trained with MIL-based BCE loss for 30 epochs, followed by 30 epochs of co-training with the LTC module.
After that, we trained the adaptor for 30,000 iterations and froze the VADor, VE, and LLaMA.
We conducted all experiments on $4$ Nvidia L40 GPUs.
We implemented the max value scores and max margin scores in Eq~\ref{eq:1} as ~\cite{lv2021localizing,lv2023unbiased}.

\subsection{Quantitative Results}
\label{sec:4.3}
\noindent\textbf{Weakly Supervised Video Anomaly Detection (WSVAD)}.
In Table~\ref{tab:ucf-crime}, we compared our VADor with other state-of-the-art (SOTA) methods in WSVAD. 
On UCF-Crime~\cite{sultani2018real}, VADor with LTC achieves the best $\mathrm{AUC}_O$ and $\mathrm{AUC}_A$ among all the methods, with an improvement of +0.88\% and +2.44\%, respectively. 
VADor with LTC achieves the second best $\mathrm{AUC}_O$ in TAD~\cite{lv2021localizing} and significantly outperforms all methods on $\mathrm{AUC}_A$ by +3.21\%.
Moreover, with the introduction of LSTC, we witness a further improvement among the two benchmarks.

\begin{table}[t]
    \centering
    \scalebox{0.9}{
    \begin{tabular}{@{}c|c|c|c}
      \toprule\hline
        Category & Method & $\mathrm{AUC}_O$ (\%) & $\mathrm{AUC}_A$ (\%)\\ 
      \hline\hline
      \multirow{8}{*}{2-Stage}
      & Sultani et al.~\cite{sultani2018real} & 75.41 &54.25    \\
      & Zhang et al.~\cite{zhang2019temporal}            & 78.66  & -    \\
      & Motion-Aware~\cite{zhu2019motion} & 79.10   & 62.18    \\
      & GCN-Anomaly~\cite{zhong2019graph} & 82.12  & 59.02    \\
      & Wu et al.~\cite{Wu2020not} & 82.44  & -    \\
      & RTFM~\cite{tian2021weakly}          & 84.30 & -   \\ 
      & WSAL~\cite{lv2021localizing}          & 85.38  & 67.38\\ 
      & ECUPL~\cite{zhang2023exploiting}          & 86.22  & -\\ \hline
      \multirow{3}{*}{E2E}
      & UMIL~\cite{lv2023unbiased}          & 86.75  & 68.68\\
      & \cellcolor{mygray}VADor w/o LTC & \cellcolor{mygray}85.90  & \cellcolor{mygray}66.67 \\ 
      & \cellcolor{mygray}VADor w LTC  & \cellcolor{mygray}\underline{87.63}  & \cellcolor{mygray}\underline{71.12} \\ 
      & \cellcolor{mygray}\textbf{VADor w LSTC}  & \cellcolor{mygray}\textbf{88.13}  & \cellcolor{mygray}\textbf{72.54} \\ \hline\bottomrule
    \end{tabular}%
    }  
    \caption{WSVAD comparison on UCF-Crime. ``$2$-stage'' and ``E$2$E'' stand for the two-stage pipeline and the end-to-end framework. ``w'' and ``w/o'' are abbreviations for ``with'' and ``without''. $\mathrm{AUC}_O$ and $\mathrm{AUC}_A$ denote that the AUC computed on the overall test set and only abnormal test videos, respectively. The best results are in bold, and the second-best results are underlined.} 
    \label{tab:ucf-crime}
    \vspace{-2mm}
\end{table}
\begin{table}[t]
    \centering
    \scalebox{0.9}{
    \begin{tabular}{@{}c|c|c|c}
      \toprule\hline
        Category & Method & $\mathrm{AUC}_O$ (\%) & $\mathrm{AUC}_A$ (\%)\\ 
      \hline\hline
      \multirow{6}{*}{2-Stage}
      & Sultani~\etal~\cite{sultani2018real} & 81.42 & 55.97    \\
      & Motion-Aware~\cite{zhu2019motion} & 83.08  & 56.89    \\
      & GIG~\cite{lv2020global}          & 85.64 & 58.65   \\ 
      & RTFM~\cite{tian2021weakly}          & 89.61  & -\\ 
      & WSAL~\cite{lv2021localizing}          & 89.64  & 61.66\\
      & ECUPL~\cite{zhang2023exploiting}          & 91.66  & -\\ \hline
      \multirow{3}{*}{E2E}
      & UMIL~\cite{lv2023unbiased}          & \textbf{92.93}  & 65.82\\ 
      & \cellcolor{mygray}VADor w/o LTC & \cellcolor{mygray}85.20 & \cellcolor{mygray}58.19 \\ 
      & \cellcolor{mygray}VADor w LTC & \cellcolor{mygray}{90.91} & \cellcolor{mygray}{\underline{69.03}} \\ 
      & \cellcolor{mygray}\textbf{VADor w LSTC} & \cellcolor{mygray}{\underline{91.77}} & \cellcolor{mygray}{\textbf{70.78}} \\ \hline\bottomrule
    \end{tabular}%
    }
    \caption{WSVAD comparison on TAD benchmark.} 
    \vspace{-2mm}
    \label{tab:tad}
\end{table}
\noindent\textbf{Overall Observations}.
1) Notice that our baseline VADor performs far better than the previous MIL-based two-stage model~\cite{sultani2018real}. This validates the strong video representation power of the pre-trained VE~\cite{zhang2023video}. 
2) Moreover, our VADor with LTC module significantly improves the $\mathrm{AUC}_A$ over VADor baseline (\eg, +4.45\% on UCF and +10.84\% on TAD), which demonstrates the effectiveness of incorporating long-range contextual information into the anomaly analysis. Additionally, the incorporation of short-term historical information leads to a further enhancement of the AUC performance.
3) Our VADor achieves the second best $\mathrm{AUC}_O$ results on TAD, it is mainly because we froze the pre-trained VE, while UMIL~\cite{lv2023unbiased} fine-tuned the feature backbone with VAD data.
The higher $\mathrm{AUC}_O$, but lower $\mathrm{AUC}_A$ of UMIL demonstrate that UMIL is better at distinguishing the normal clips in normal videos, but our VADor achieves a better anomaly localization performance among anomalous videos with a much higher $\mathrm{AUC}_A$ (+3.21\%).
\begin{figure}[t]
    \centering
    \footnotesize
    \includegraphics[width=1\columnwidth]{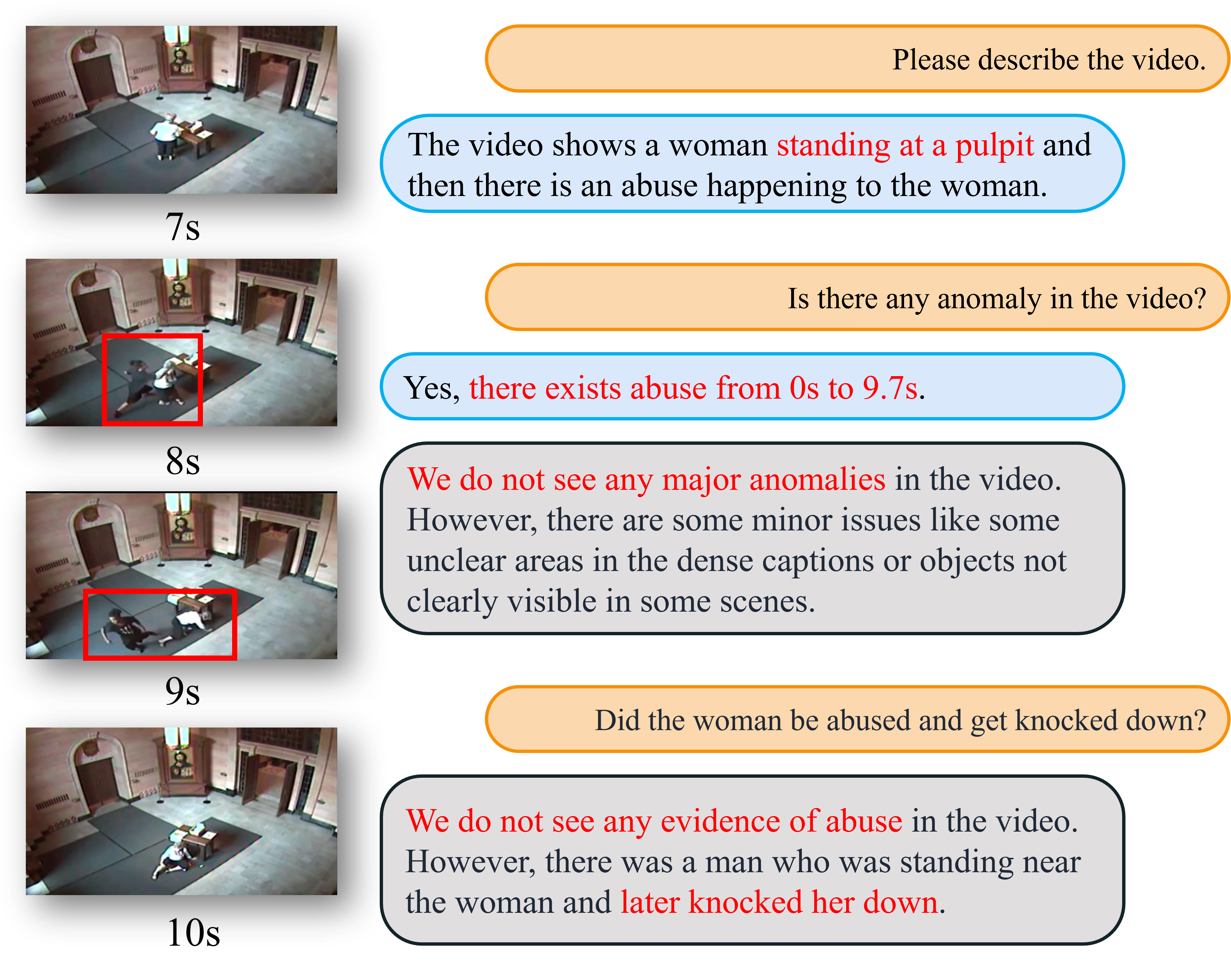}
	\caption{An abuse example for comparison between the VAD-LLaMA and Video-LLaMA. The \textcolor{red}{red} boxes in the frames are ground-truth anomalies. The \textcolor{orange}{orange} boxes are the question from humans. The \textcolor{gray}{gray} and \textcolor{iconblue}{blue} boxes are the answers from the Video-LLaMA and our VAD-LLaMA, respectively. Best viewed in color.}
    \vspace{-2mm}
	\label{fig:vadcase_1}
\end{figure}

\subsection{Quantitative Examples}
\begin{figure}[t]
	\centering
    \begin{subfigure}[t]{0.45\textwidth}
         \includegraphics[width=\textwidth]{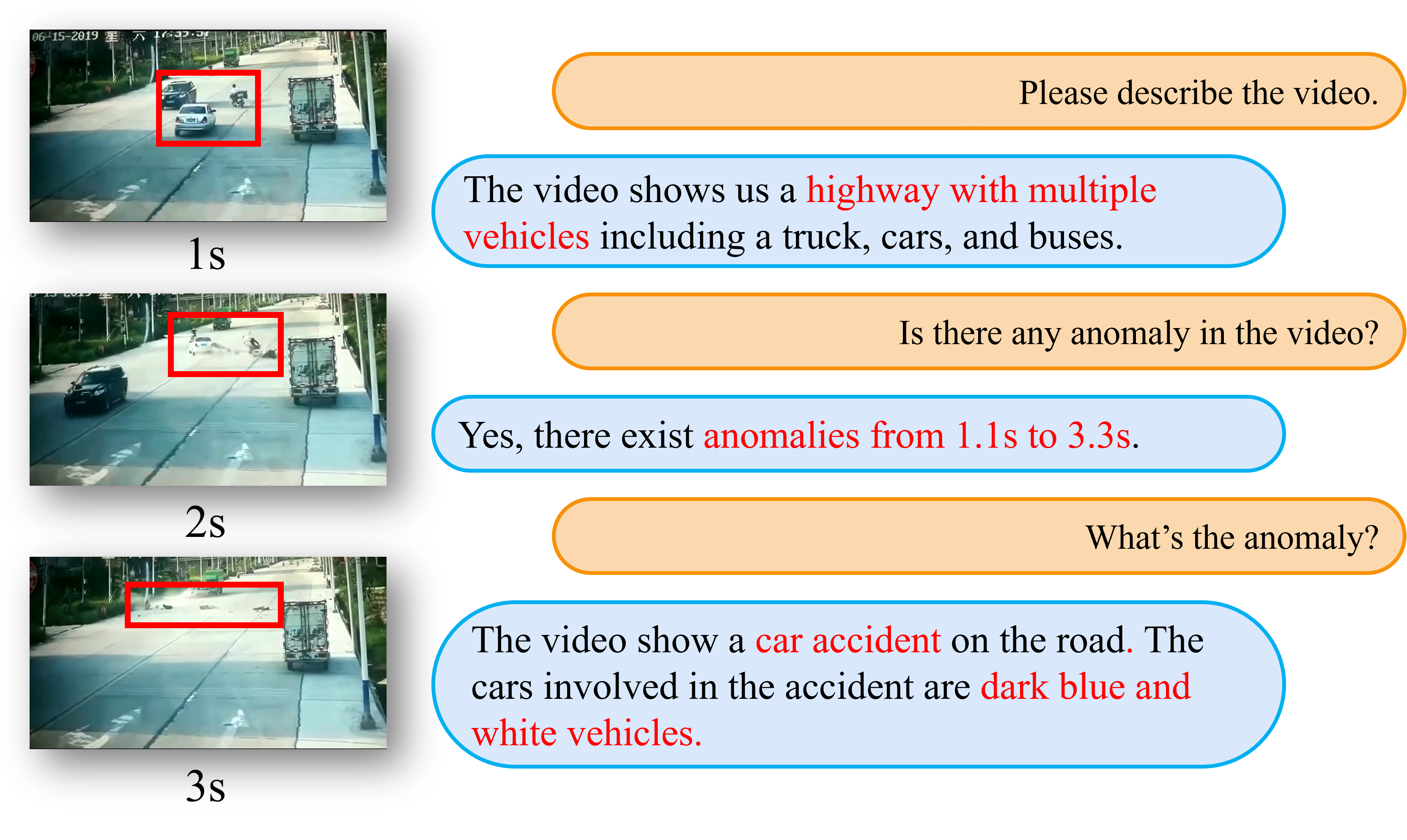}
         \caption{Car accident}
         \label{fig:1a}
    \end{subfigure}
    \begin{subfigure}[t]{0.45\textwidth} 
         \includegraphics[width=\textwidth]{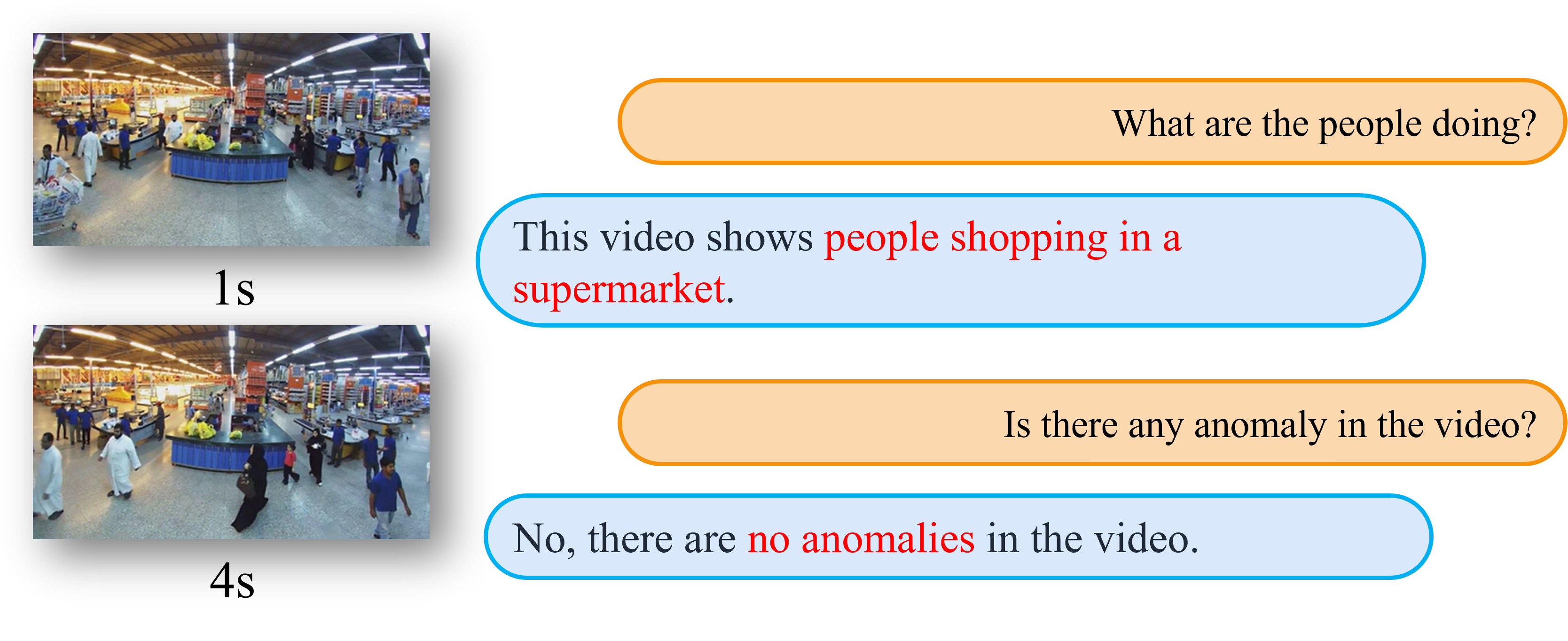}
         \caption{Normal}
         \label{fig:1b}   
    \end{subfigure}
	\caption{Two qualitative examples of VAD-LLaMA.}
	\label{fig:vadcase_2}
    \vspace{-2mm}
\end{figure}

We showcase a video example of `Abuse' for comparison between our VAD-LLaMA and Video-LLaMA~\cite{zhang2023video} in Figure~\ref{fig:vadcase_1}.
As observed, the Video-LLaMA, available as an open-source model, struggles to precisely correlate the detected anomaly (Abuse) with specific events in the video, notably the incident where a woman is knocked down.
Additional examples of our model with examples are put in Figure~\ref{fig:vadcase_2}. The model effectively identifies anomalies, \ie, ``Abuse'' and ``Car accident'', accurately pinpoints their temporal locations, and provides a detailed description of the anomalies.
For normal videos, our VAD-LLaMA is able to comprehensively analyze the content of the video and eliminate the possibility of anomalies. Moreover, users are able to engage in multi-turn dialogues pertaining to the video content. More qualitative examples and comparisons between VAD-LLaMA and Video-LLaMA are moved to the Appendix.

\subsection{Ablation studies}
\label{sec:Abla}
\noindent\textbf{LTC Components}. 
We validate the effectiveness of long-term context modeling in Table~\ref{tab:ablation} with $\mathrm{AUC}_O$. 
By comparing the third and fourth lines with the first line, we observe that the normal (abnormal) features in long-range video context can improve $\mathrm{AUC}_O$ from 85.90\% to 87.08\% (87.45\%) on UCF-crime and 85.20\% to 88.39\% (89.08\%) on TAD. 
In the fifth line, with the combination of the normal and abnormal contexts, a further improvement of $\mathrm{AUC}_O$ proves the critical role of the long-range video context in robust anomaly mining.
In addition, we verify the success of short-term historical information in VADor, which boosts the $\mathrm{AUC}_O$ to reach 88.13\% on UCF-crime and 91.77\% on TAD.
For an independent evaluation of the effectiveness of our VADor, we re-implemented the previous SOTA UMIL~\cite{lv2023unbiased} using the VE features, denoted as UMIL*. The results are presented in line 2. Our VADor with LTC in line 4 consistently outperforms UMIL* (+0.85\% on UCF-Crime and +2.26\% on TAD), thereby validating the efficacy of our design, based on the same feature backbone.

\begin{table}[t]
\centering
\scalebox{1.0}{
\begin{tabular}{ccccc|cc}
\toprule\hline
Baseline & Nor & Abn & His & UMIL & UCF & TAD \\ \hline \hline
\checkmark & & & & & 85.90 &  85.20\\
\checkmark & & & & \checkmark & 86.78 & 88.65 \\ \hline
\checkmark & \checkmark & & & & 87.08 & 88.39 \\ 
\checkmark & & \checkmark & & & 87.45 & 89.08 \\ 
\checkmark & \checkmark & \checkmark & & & 87.63 & 90.91 \\ 
\cellcolor{mygray}\checkmark & \cellcolor{mygray}\checkmark & \cellcolor{mygray}\checkmark & \cellcolor{mygray}\checkmark & \cellcolor{mygray} & \cellcolor{mygray}\textbf{88.13} & \cellcolor{mygray}\textbf{91.77} \\ \hline
\bottomrule
\end{tabular}%
}
\caption{Ablation studies of the components in the LTC module. Here, ``Nor'' and ``Abn'' denote the normal and abnormal list, respectively. ``His'' stands for the short-term history list and ``UMIL'' is the unbiased term proposed in~\cite{lv2023unbiased}. }
\vspace{-3mm}
\label{tab:ablation}
\end{table}
\noindent\textbf{LTC Length}. 
In the LTC, $K$ is employed as the length of the feature lists. Through empirical analysis presented in Table~\ref{tab:k}, we determine that $K=4$ is a suitable choice across the two datasets, and thus, it is the default setting for our experiments. 
In general, the selection of $K$ hinges on the reliance of anomaly detection on video contexts. For instance, a small $K$ might not capture sufficient temporal information, while a large $K$ could involve unexpected noise.

\begin{figure}[t]
	\centering
	\includegraphics[width=\linewidth]{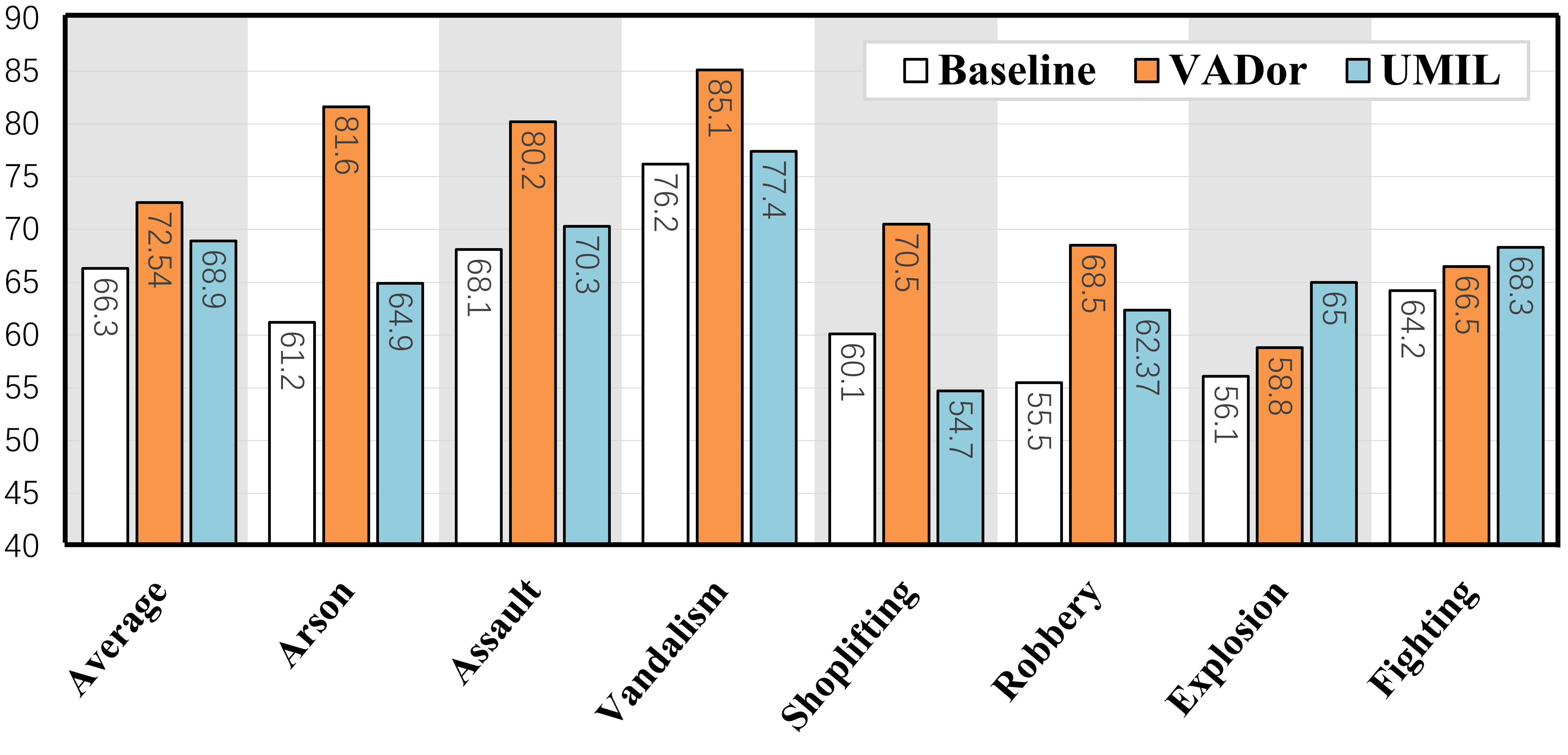}
	\caption{Class-wise $\mathrm{AUC}_A$ of three methods on UCF-Crime. Here, ``VADor'' stands for our VADor with the LTC module.}
	\label{fig:hist}
    \vspace{-2mm}
\end{figure}

\noindent\textbf{Class-wise AUC}. 
On the UCF-Crime dataset, each test video is labeled with the class of anomaly, enabling us to analyze models' capabilities in detecting subtle abnormal events through class-wise $\mathrm{AUC}_A$ comparisons. In Figure~\ref{fig:hist}, we compare VADor with the baseline and UMIL, where ``Average" represents the overall $\mathrm{AUC}_A$, and the remaining bars show the class-wise values.

Our observations are as follows:
1) Both the VADor baseline and UMIL demonstrate strong performance on anomaly classes characterized by drastic motions, such as ``Assault" and ``Vandalism". These classes represent intuitive anomalies primarily relying on feature representation learning in a short time duration, given that VE and the backbone of UMIL are sufficient to capture local details in short video clips.
2) However, these methods struggle to distinguish anomalies that depend on long-range temporal analysis, like ``Arson" and ``Shoplifting". These classes correspond to the hard examples, which are inadequately addressed by the long-term context modeling in our VADor.
VADor with the LTC module performs similarly well on the aforementioned intuitive anomaly classes and significantly outperforms the other methods on other anomalies that require comprehensive context modeling. This substantial improvement contributes to the superior anomaly detection performance.
Overall, observations 1 and 2 empirically validate the effectiveness of mining long-range video contexts for a more robust anomaly analysis.

\begin{table}[t]
\centering
\scalebox{0.9}{
\begin{tabular}{cccccc}
\toprule\hline
Threshold(\%) & 0 & 2 & \cellcolor{mygray}\textbf{4} & 6 & 8  \\ \hline \hline
$\mathrm{AUC}_O$ (\%) - UCF & 85.90 & 87.49 & \cellcolor{mygray}\textbf{87.63} & 87.27 & 87.18 \\ 
$\mathrm{AUC}_O$ (\%) - TAD & 85.20 & 90.18 & \cellcolor{mygray}\textbf{90.91} & 90.65 & 90.13 \\ \hline
\bottomrule
\end{tabular}%
}
\caption{Ablation of the LTC Length on UCF-Crime and TAD.}
\vspace{-3mm}
\label{tab:k}
\end{table}



\section{Conclusion}
\label{sec:Con}
In this work, we introduced VAD-LLaMA, a novel Video Anomaly Detection (VAD) approach that integrates video-based large language models (VLLMs) into the VAD framework, making the VAD model free from thresholds and able to explain the reasons for the detected anomalies. 
In our model, we introduced a Long-Term Context (LTC) module to mitigate the incapability of existing VLLMs in long-range context modeling.
In addition, our three-phase training method significantly improves the efficiency of training VLLMs in specific domain as VAD by minimizing the requirements for VAD data and reducing the costs of annotating instruction-tuning data. 
Our approach was empirically validated by the state-of-the-art performance and extensive ablations on standard WSVAD benchmarks. 
Also, we showcased the anomaly localization and description capability of VAD-LLaMA in the multi-dialogue based on the video content.
In the future, we seek to develop a VAD model with fast adaption capability that can detect new anomalies based on either a few example clips or textual descriptions of the targeted anomalies.

{\small
\bibliographystyle{ieee_fullname}
\bibliography{egbib}
}

\end{document}